\documentclass[10pt,twocolumn,letterpaper]{article}

\usepackage{cvpr}              

\usepackage{tikz}
\usetikzlibrary{positioning}
\usepackage{subcaption}
\usepackage{graphicx}
\usepackage{multirow}
\usepackage{tabularx}
\usepackage{bbding}
\usepackage{xcolor}
\usepackage{colortbl}
\usepackage{makecell}
\usepackage[table,xcdraw]{xcolor}
\usepackage{minted}
\usepackage{appendix}

\usepackage{arydshln}
\definecolor{grey}{rgb}{0.9, 0.9, 0.9}

\newcommand{\ours}{\textit{{Ego}}\xspace}
\newcommand{\bX}{\mathbf{X}}
\newcommand{\bV}{\mathbf{V}}
\newcommand{\bK}{\mathbf{K}}
\newcommand{\bQ}{\mathbf{Q}}

\newcommand{\bA}{\mathbf{A}}

\definecolor{cvprblue}{rgb}{0.21,0.49,0.74}
\definecolor{singleconcept1}{RGB}{230,245,255} 
\definecolor{multiconcept1}{RGB}{255,240,230} 
\definecolor{singleconcept2}{RGB}{220,235,245} 
\definecolor{multiconcept2}{RGB}{245,230,220} 
\definecolor{singleconcept3}{RGB}{210,225,235} 
\definecolor{multiconcept3}{RGB}{235,220,210} 
\definecolor{videocolor}{RGB}{210,240,200} 

\newcommand{\myparagraph}[1]{\noindent\textbf{#1}}
\usepackage[pagebackref,breaklinks,colorlinks,allcolors=cvprblue]{hyperref}

\def\paperID{10085} 
\def\confName{CVPR}
\def\confYear{2026}

\title{Ego: Embedding-Guided Personalization of Vision-Language Models}

\author{
Soroush Seifi\thanks{Providing contracted services at Toyota Motor Europe.} \quad
Simon Gardier \quad
Vaggelis Dorovatas\textsuperscript{*} \quad
Daniel Olmeda Reino \quad
Rahaf Aljundi\\[4pt]
Toyota Motor Europe \\
}

\begin{document}
\maketitle
\begin{abstract}
AI assistants that support humans in daily life are becoming increasingly feasible, driven by the rapid advancements in multimodal language models. A key challenge lies in overcoming the generic nature of these models to deliver personalized experiences. Existing approaches to personalizing large vision language models often rely on additional training stages, which limit generality and scalability, or on engineered pipelines with external pre-trained modules, which hinder deployment efficiency.
In this work, we propose an efficient personalization method that leverages the model’s inherent ability to capture personalized concepts. Specifically, we extract visual tokens that predominantly represent the target concept by utilizing the model’s internal attention mechanisms. These tokens serve as a memory of that specific concept, enabling the model to recall and describe it when it appears in test images.
We conduct a comprehensive and unified evaluation of our approach and  SOTA  methods across various personalization settings including single-concept, multi-concept, and video personalization,  demonstrating strong performance gains with minimal personalization overhead.
\end{abstract}    
\section{Introduction}
\label{sec:intro}

\begin{figure}[htbp]
    \centering
      \includegraphics[width=\columnwidth]{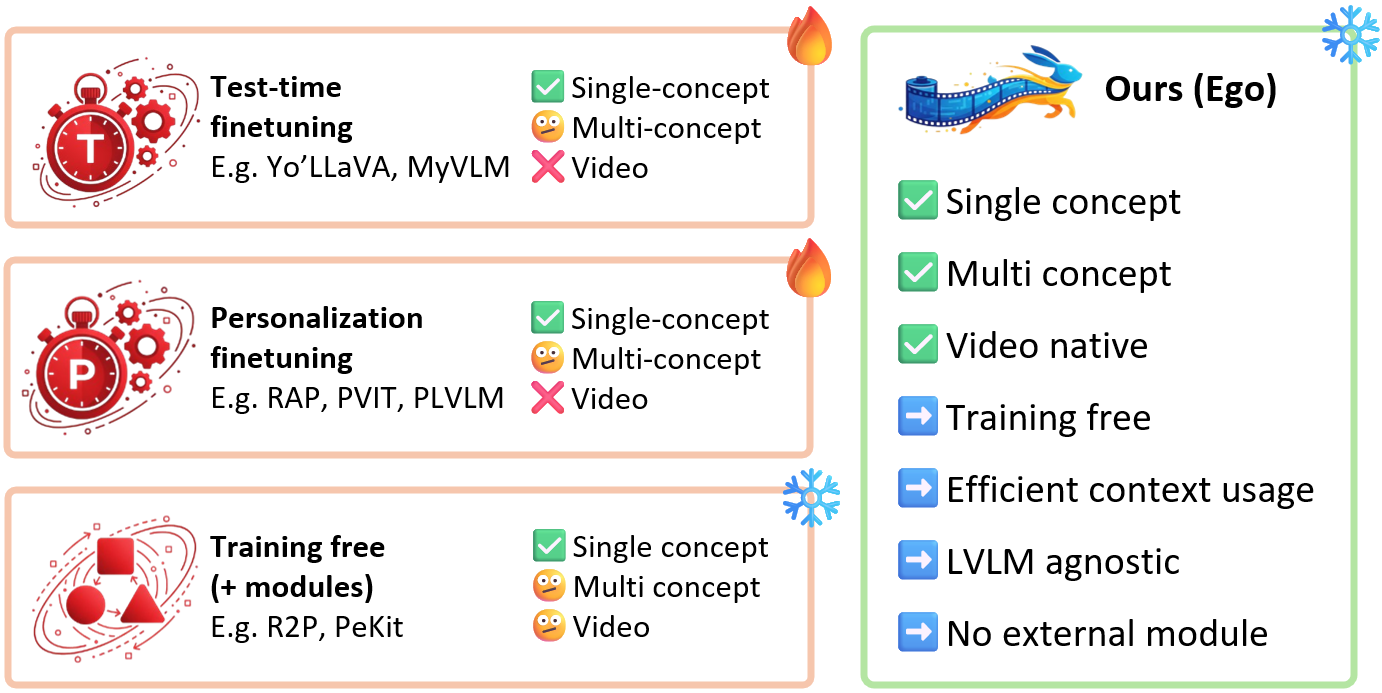}
      \caption{
      \textbf{Personalization approaches vs \ours}. Existing methods typically require test-time or LVLM fine-tuning, or depend on external vision modules, and often fail to support multi-concept or video-level personalization. In contrast, \ours is training-free, LVLM-agnostic, requires no external modules, and efficiently enables single-concept, multi-concept, and video-native personalization within a unified framework.
      }
      \vspace{-0.3cm}
      \label{fig:teaser-methods-characteristics}
    \end{figure}

\begin{figure*}[htbp]
      \centering
      \includegraphics[width=0.8\textwidth]{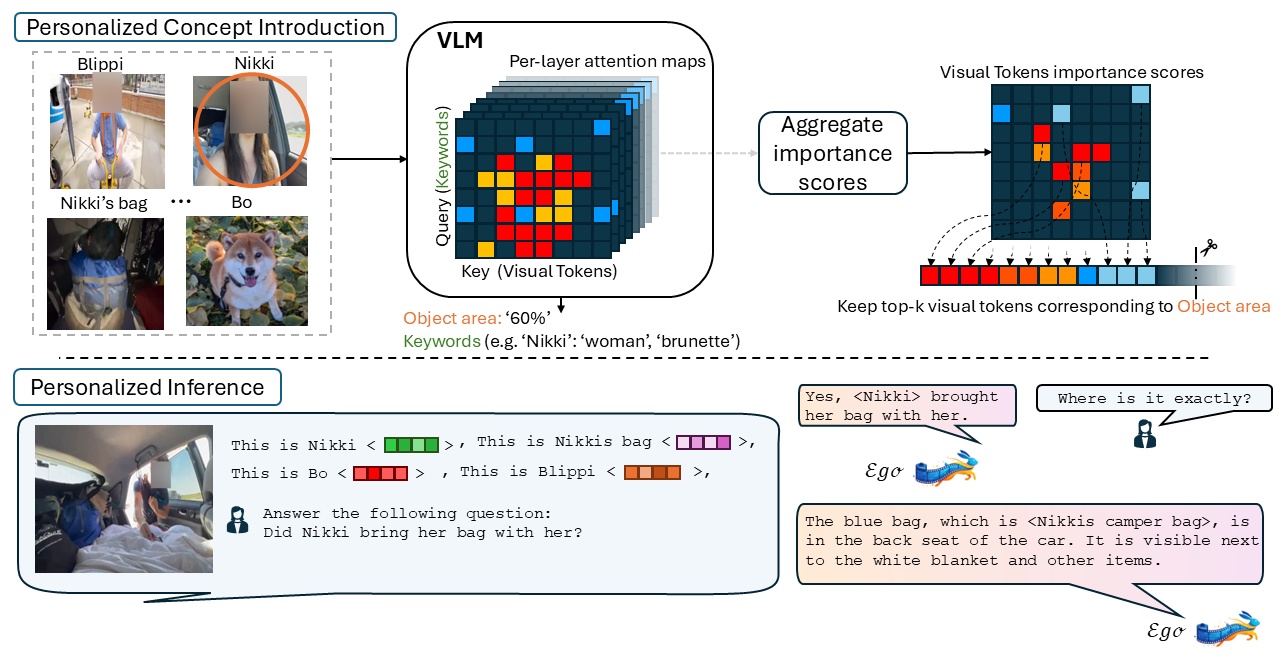}
      \vspace{-0.2cm}
      \caption{\textbf{Our proposed method \ours}.  \textbf{Personalized Concept Introduction:} The LVLM is tasked to estimate the subject area in the reference image and generate keywords describing its main characteristics.  \ours  identifies the most representative visual tokens via keywords cross cross-attention and creates a concept memory. \textbf{Inference:} Given a test image, the LVLM in \ours  accesses internal concept memories \textbf{in context} to recall and reason about known subjects in the image. \ours requires neither additional training nor external modules.}
      \label{fig:teaser-method}
      \vspace{-0.3cm}
\end{figure*}

Large Language Models (LLMs) have recently demonstrated impressive capabilities in understanding, reasoning, and generating text across diverse domains~\cite{mann2020language, singhal2023large, schick2023toolformer}. Extending these capabilities to the multimodal domain, Large Vision-Language Models (LVLMs) have achieved notable success in tasks such as image captioning~\cite{radford2021learning}, visual question answering~\cite{li2023blip, liu2023visual}, and embodied navigation~\cite{shah2023lm}. These advances position LVLMs as promising general-purpose perceptual agents. As human-AI interaction becomes increasingly multimodal, the ability of LVLMs to detect, understand, and reason about individual users and their belongings is essential for personalized experience.

Personalization of Large Vision-Language Models seeks to adapt pre-trained models to recognize, describe, and reason about user-specific entities~\cite{alaluf2024myvlm, nguyen2024yo}. Unlike general-purpose LVLMs that operate at the category level, personalization focuses on capturing the unique visual and semantic traits of specific subjects—such as a person, object, or pet—using limited reference data. This capability unlocks a wide range of applications, including personalized text generation, user-specific assistants, and long-term embodied agents that maintain consistent knowledge of users and their environments. More broadly, effective personalization bridges the gap between general visual-language understanding and individualized, context-aware intelligence.

Despite increasing interest in personalization, current approaches face several practical limitations. Many methods rely on test-time finetuning for each individual subject~\cite{alaluf2024myvlm, nguyen2024yo}, which significantly hampers scalability, especially on resource-constrained edge devices. Some recent works attempt to bypass test-time finetuning by leveraging large-scale training and instruction tuning of LVLMs to generate personalized outputs~\cite{pi2024personalized, hao2025rap}. However, even after training for personalization, these models typically require reference views of the personalized concepts during inference~\cite{pi2024personalized, hao2025rap, pham2024personalized}, adding computational overhead and forcing the model to reprocess the personalized input each time. On the other hand, training-free approaches often depend on heavily engineered architectures or external vision modules~\cite{das2025training, seifi2025personalization}, leading to increased system complexity and inference-time overhead, refer to Fig.\ref{fig:teaser-methods-characteristics} for an illustration.

Leveraging the strong image understanding and in-context learning capabilities of modern LVLMs~\cite{kang2025online, chen2025true}, our method eliminates the need for additional training, fine-tuning, or external tools. We address personalization by enabling the model to build an internal memory of personalized concepts. Upon introduction of a concept via one or a few reference views, we task the model to describe its distinctive features. During this process, we extract the most attended visual tokens—those the model deems most representative—and aggregate them into a compact concept visual memory.
At inference, these concepts memories are provided in context, guiding the model to recognize and personalize its response when the subject reappears, Fig.~\ref{fig:teaser-method} illustrates our approach.


We term our method {\ours}, short for Embedding-Guided Personalization of Vision-Language Models. {\ours} operates entirely without additional training, external modules, while minimizing inference-time overhead. 

In comparing our approach with existing state-of-the-art methods, we identified substantial inconsistencies in the datasets and evaluation protocols employed across prior work. To ensure fairness and reproducibility, we perform a unified evaluation of representative personalization techniques on diverse datasets and tasks, including recognition, visual question answering, and captioning. Our evaluation spans multiple personalization scenarios, covering single-concept, multi-concept, and video personalization where applicable. \ours achieves the strongest training-free performance in single-concept personalization and delivers significant improvements over both training-based and non-training approaches in multi-concept and video personalization.


Our key contributions are as follows:
1)  We propose \ours, a training-free personalization method that requires no fine-tuning, external tools, or architectural changes.
2) We conduct a unified evaluation of state-of-the-art approaches  establishing a comprehensive testbed for future works.
3)  {\ours} achieves state-of-the-art performance with  minimal compute overhead.
4) {\ours} supports single-concept, multi-concept, and video personalization within a unified model-generic framework.



\section{Related Work}
\label{sec:formatting}
Personalization of pretrained models refers to adapting the behavior of generic models to fit specific user profiles, concepts, or styles, and it has attracted growing interest across multiple domains. Large Language Model (LLM) personalization focuses on tailoring models to user or group preferences derived from past interactions or explicit feedback (e.g.,~\cite{zhang2024personalization, lai2024large, wu2024exploring}), while image and video generation personalization enables integrating specific concepts into generated media (e.g.,~\cite{ruiz2023dreambooth, ruiz2024hyperdreambooth}). Encoder-based vision-language personalization typically addresses concept retrieval (e.g.,~\cite{gal2022image, cohen2022my}). In this work, we focus on Large Vision-Language Model (LVLM) personalization, which aims to enhance LVLMs’ ability to understand, reason, and respond to visual input by incorporating knowledge about personalized concepts such as individuals and their belongings. Current LVLM personalization approaches can be broadly categorized into the following lines of work. 




\myparagraph{Test-Time Fine-Tuning.}  
Early attempts to LVLM personalization approach the task via a dedicated training phase for each introduced concept. \textbf{MyVLM}~\cite{alaluf2024myvlm} learns binary classification heads for each personalized concept and injects concept identifiers through a Q-former mechanism, while  
\textbf{Yo’LLaVA}~\cite{nguyen2024yo} employs prompt-tuning with trainable prefix tokens per concept and fine-tunes LLM classifier weights. 
Both rely on caption-level supervision and QA pairs, focusing   on object-centric evaluation without addressing multi-concept or video personalization~\cite{seifi2025personalization}.

 \myparagraph{Finetuning for Personalization.}  
Training approaches remove test-time tuning requirement by enabling models to recognize personalized concepts directly from reference images~\cite{hao2025rap, pi2024personalized, pham2024personalized}.  
\textbf{PLVLM}~\cite{pham2024personalized} trains an aligner module using DINOv2 embeddings but focuses on human-centric evaluation without exploring generalization to other categories.  
\textbf{PVIT}~\cite{pi2024personalized} builds a synthetic dialogue dataset for personalized conversations and fully fine-tunes the LVLM on it.  
\textbf{RAP}~\cite{hao2025rap} retrieves candidate concepts via visual similarity and uses LoRA-based fine-tuning on large scale paired data.
Although these methods avoid per-concept tuning, they remain resource-intensive and rely on reference views at inference, which can introduce context-length bottlenecks. Furthermore, training biases the model towards the constructed personalization paired data, limiting scalability to multi-concept setting, as shown in our experiments.

\myparagraph{Training-Free Methods:}  
As LVLMs become more powerful, training-free methods aim to enable personalization without altering the LVLM, prioritizing   efficiency and scalability.
\textbf{R2P}~\cite{das2025training} generates descriptive attributes per concept and detects them via top-$k$ retrieval using external vision models~\citep{johnson2019billion,radford2021learning}.
\textbf{PeKit}~\cite{seifi2025personalization} decouples object detection from LVLM reasoning using an external segmentation network and a DINOv2-based memory bank. 
These methods still depend on external modules and trade off training for additional test-time computation. In contrast, our approach, \ours, tackles personalization by leveraging the LVLM’s strong visual understanding capabilities. We create visual concept memories that compress subject attributes and are accessed in context by the LLM, requiring minimal  overhead equivalent to textual prompting at inference.

\section{Ego: Embedding-Guided Personalization }
\label{sec:method}
Our approach is motivated by the observation that recent powerful LVLMs are capable of cross-referencing objects in multiple input images and reasoning about objects in short videos~\cite{zhu2025internvl3,bai2025qwen2}. This shows an inherent ability to recognize objects across different images or video frames, implying that the model internally assigns discriminative embeddings to individual objects for recognition and tracking.

Leveraging this insight, we want to extract these discriminative embeddings from the LVLM’s intermediate representations and provide them as in-context information, enabling the model to identify and reason about personalized concepts. The proposed method is modular, scalable to multiple concepts and video inputs, and compatible with any number of reference views. The following sections detail each component of our approach.
\subsection{Preliminaries}
Given a Large Vision Language model $\mathcal{M}$ composed of an LLM, and a visual encoder with a projector (collectively referred to as {VP}) that maps input image(s) into the LLM embedding space referred  to hereafter as~\textbf{ visual tokens}. Our objective is to enable $\mathcal{M}$ to generate textual outputs tailored to personalized concepts. For a concept $c$ to be added to the personalized concept set $C$, we receive a set of reference images  $\{R_c\}$ along with its concept name $n_c$.  Our method works with one or multiple reference images.
For each reference view $R_c$ of concept $c$,  we query $\mathcal{M}$ with an instruction $I$ to produce key descriptive words. Formally:
\begin{equation}
T= \mathcal{M}(R_c, I); \, T= \text{LLM}(\mathbf{X}_R, I), \, \text{where } \mathbf{X}_R = \text{VP}(R_c),
\end{equation}
$\mathbf{X}_R \in \mathbb{R}^{N_r\times D}$,  $N_r$ is the number of visual tokens extracted from a reference  image and $D$ is the LLM token embedding dimension, and $T$ is the textual output of the LLM. From $T$ we filter out punctuations and keep textual tokens corresponding to the LLM output keywords $\mathbf{W}$. Our objective is to select a compact subset of visual tokens that best represent the personalized subject in the given reference image. This is motivated by recent works highlighting redundancy in visual tokens and showing that selecting an informative subset increases efficiency and can often outperform using the full image or video~\citep{dorovatasrecurrent, huang2025llmvtp, chen2024image}.

\subsection{Attention-Guided Embedding Extraction }
A reference image $R_c$, mapped into visual tokens $\mathbf{X}_R $ by $VP$, contains the subject and unrelated background. We aim to identify and extract a minimal subset of visual tokens $\mathbf{X}_R^c \in \mathbb{R}^{K\times D}$ (where $K  << N_r$ ) that captures the unique characteristics of a given concept $c$. This approach serves two main purposes. First, efficiency: representative tokens can be aggregated from multiple reference views while keeping the overall token count per concept manageable for personalized inference. Second, relevance: it is essential to remove tokens that do not represent the subject, such as those corresponding to background elements, so the concept representation remains focused and accurate.

To identify concept-specific embeddings, we analyze the attention maps of the LLM layers, specifically, we focus on cross-modal attention of the keywords tokens $\mathbf{W}$ to the visual tokens $\mathbf{X}_R $. We hypothesize that the representative visual tokens corresponding to the  descriptive keywords will receive the highest attention scores by the keywords embeddings and, the higher the attention score, the higher the importance of the visual token. 
 First, we describe how to compute importance scores from attention maps per visual token given all LLM attention heads and layers, then we explain how we identify the most relevant layers for visual objects understanding in a given $\mathcal{M}$.
 
 For a layer $l$ and attention head $h$, the full attention matrix is computed on the embedding $\mathbf{X}^l$ after mapping it to Query $\bQ^{l,h},$ Key $\bK^{l,h} $ {and Value}  $\bV^{l,h}  $
 \begin{equation}
    \bA^{l,h} = \text{Softmax} \left( \frac{\bQ^{l,h} (\bK^{l,h})^\top}{\sqrt{d_k}} \right),
\end{equation}

where $d_k$ is the dimension of attention heads.
We extract the  cross attention matrix $A_{wr}^{l,h} \in \mathbb{R}^{ N_w \times N_r }$ computed on the  embedding at layer $l$, $\mathbf{X}^l=[\mathbf{X}^l_R,\bX^l_W]$ where $N_w$ is the number of tokens in the key descriptive words. 
The extracted $A_{wr}^{l,h}$  represents the cross attention scores between the keywords embedding  $\bX^l_W$ and the visual tokens $\mathbf{X}^l_R$.



Given a set of relevant LLM layers $L$, we compute an importance score for each visual token  $\mathbf{X}^l_R[j]$ by maximizing attention scores over heads and layers and averaging the scores from the different keywords tokens . 
\begin{equation}
I_{j} = \frac{1}{|L|} \sum_{l \in L}^{} \frac{1}{H}\sum_{h=1}^{H} \left( \frac{1}{N_w} \sum_{n=1}^{N_w} {\bA^{l,h}_{wr}}[n,j] \right).
\end{equation}

With an importance score per visual token, we select  the  $K_c$ most important  visual tokens of $\bX_R$ as representative of the concept introduced for personalization. Specifically:
\[
\begin{aligned}
\mathcal{P}^{\text{ordered}}
&=
\overbrace{\operatorname{sort}_{\uparrow}\!}^{\text{restore order}}\Big(\overbrace{
    \operatorname{argsort}_{\downarrow}(\mathbf{I})[1:K_c]}^{\text{select top-$K_C$}}
\Big), \\[6pt]
\mathbf{X}_R^{c}
&= \mathbf{X}_R[\mathcal{P}^{\text{ordered}}, :].
\end{aligned}
\]



As noted before, we select a subset of the visual tokens $\mathbf{X}_R^c \in \mathbb{R}^{K_c \times D}$ that received the highest attention. When provided with multiple reference views $N_v$, we process each image independently to extract the top visual tokens which will be concatenated  into one matrix: $\mathbf{X}_R^c\in \mathbb{R}^{N_v*K_c \times D}$. We construct $\mathbf{X}_R^c$  to represent the LVLM memory of the personalized concept and its most unique visual characteristics, acting as the visual highlights of a given subject. During inference, the model receives each concept’s visual memory $\mathbf{X}_R^c$ along with its name $n_c$, and is tasked with determining whether the personalized concept(s) appear in the inference image and respond to the input query accordingly.

\subsection{Concept Memory Size}\label{sec:memsize}
We select a small set of visual tokens, denoted as $K_c$, to capture the model’s memory of the personalized concept. The optimal number of representative tokens depends on the size of the object in the reference image. For instance, when the object occupies a small area (e.g., a phone or a pair of shoes), only a few tokens are sufficient. Conversely, for larger subjects such as a person in a high-resolution profile image, a greater number of tokens is preferred.

To build a compact memory of a personalized subject while considering the area it occupies in the image, we leverage the LVLM’s inherent ability to estimate the size of the region taken up by the main subject in the reference image.
Given a reference image $R_c$ of a newly introduced concept $c$, we first ask the LVLM to estimate the percentage of the area that the subject occupied in the image ($\alpha_c$) and then estimate the appropriate number of visual tokens as
  $K_c=min(K, \frac{\alpha_c\times N_r}{100})$, 
where  $K$ is the maximum number of visual tokens allowed per reference image. Setting a maximum number $K << N_r$ is essential to maintain efficiency during inference and to extract only patches representing the key attributes of the subject. 

%

\subsection{Layer Selection}\label{sec:layerselction}
LVLMs encode information across layers with varying levels of abstraction. Our objective is to identify the layers that exhibit the strongest interaction between visual tokens and the generated keywords for a given subject. Prior work suggests that mid-to-late layers enhance visual representations~\cite{jiang2025devils}, dominate visual-to-text information flow~\cite{kim2025interpreting}, and enable efficient fusion through attention to text-relevant regions~\cite{fan-etal-2025-visipruner}.  
However, there is no established method for determining  vision-relevant layers in a specific LVLM. To address this, we propose an automatic procedure that identifies where the generated text interacts  most with the visual tokens in an image, using an external calibration set.  

We sample a subset of images from the COCO 2017 training split \cite{lin2014microsoft}, each containing a single category and one instance of that category. Using the ground-truth segmentation mask for each object instance, we determine which visual tokens correspond to the object. The LVLM is then tasked with describing the main foreground object in  each image. For every layer, we compute the overlap between the top $K$ patches (ranked by our importance score) and the segmentation mask, and rank layers based on their average overlap. The top $L$ layers are selected.  

This process yields a set of layers where image descriptions typically interact most strongly with the visual information of the main subject, serving as a proxy for our task of understanding a subject in a reference image. Note that this calibration is performed only once per LVLM.

\subsection{Ego Inference}

\ours accesses the model's internal states and attention maps to construct a model's own memory $\{\mathbf{X}_R^c, n_c\}$ of each personalized concept. At  inference, we retrieve the memories of the personalized concepts and inject them  into the context of the LLM as soft prompts. We then instruct the LVLM to check if the provided concepts are present in the image and to  respond to the query accordingly. When the number of the personalized concepts increases beyond the context limit of a given LVLM, a filtering step can be done based on the similarity of the query image (in the LLM embedding space) to the stored concepts' memories. Note that by storing the visual tokens  in the LLM embedding space, we avoid the need for reprocessing the reference view at test-time by the visual encoder and allow for a small memory and compute footprint during inference, as we show in the experiments~\ref{sec:exp}.
Our soft prompting leverages recent LVLMs capability of In Context Learning~\citep{chen2025true, baldassini2024makes}, alleviating the need for additional stages of training and alignment. 
This mechanism supports single-concept, multi-concept, and video-level personalization with a unified procedure.

\section{Experiments}\label{sec:exp}
We present a comprehensive evaluation of \textit{Ego} across diverse personalization settings. We compare its performance against both training-free and training-based SOTA methods, analyze  run-time, and conduct ablation studies to assess the impact of key design choices.

\subsection{Experimental Settings}

Existing  personalization methods vary widely in datasets and evaluation metrics~\cite{alaluf2024myvlm, nguyen2024yo, seifi2025personalization, hao2025rap}. Even when using the same dataset, evaluation splits often differ~\cite{hao2025rap, das2025training}.
To ensure fair comparison, we standardize all experimental settings—model backbones, preprocessing, and evaluation protocols—and reproduce prior results under identical conditions. Our implementation will be publicly released.

\myparagraph{Compared Methods}
We benchmark representative SOTA approaches. For finetuning-based methods, we select RAP~\cite{hao2025rap}, which learns from large-scale synthetic personalization examples across tasks. For training-free post-hoc methods, we include R2P~\cite{das2025training} and PeKit~\cite{seifi2025personalization}, both neither fine-tune nor modify the base model.
We exclude test-time finetuning methods (MyVLM~\cite{alaluf2024myvlm}, Yo’LLaVA~\cite{nguyen2024yo}) due to their impractical per-concept fine-tuning and consistent underperformance~\cite{hao2025rap,das2025training,seifi2025personalization}.

\begin{table*}[ht!]
\centering
\caption{\textbf{Recognition.} Performance comparison across methods and settings using InternVL3 (14B)~\cite{zhu2025internvl3} and Qwen2.5-VL (7B)~\cite{bai2025qwen2}. Best results are shown in \textbf{bold}, and second-best results are \underline{underlined}. For training-free methods, the Training Time column reflects the average time required for concept introduction. \ours attains state-of-the-art recognition accuracy across datasets while introducing concepts with minimal overhead. Note that RAP dataset is limited to a single-reference training set, and R2P~\cite{das2025training} does not support multi-concept tasks.}
\resizebox{\textwidth}{!}{
\begin{tabular}{llc
>{\columncolor{singleconcept1}}c
>{\columncolor{singleconcept2}}c
>{\columncolor{singleconcept3}}c
>{\columncolor{singleconcept1}}c
>{\columncolor{singleconcept2}}c
>{\columncolor{singleconcept3}}c
>{\columncolor{singleconcept1}}c
>{\columncolor{singleconcept2}}c
>{\columncolor{singleconcept3}}c
>{\columncolor{multiconcept1}}c
>{\columncolor{multiconcept2}}c
>{\columncolor{multiconcept3}}c
>{\columncolor{multiconcept1}}c
>{\columncolor{multiconcept2}}c
>{\columncolor{multiconcept3}}c}
\toprule
& & &
\multicolumn{9}{c}{\textbf{Single Concept}} &
\multicolumn{6}{c}{\textbf{Multi Concept}} \\
\cmidrule(lr){4-12} \cmidrule(lr){13-18}
\textbf{Method} & \textbf{Model} & \makecell{\textbf{Training} \\ \textbf{Time} $\downarrow$}&
\multicolumn{3}{c}{\textbf{MyVLM}~\cite{alaluf2024myvlm}} &
\multicolumn{3}{c}{\textbf{Yo’LLaVA}~\cite{nguyen2024yo}} &
\multicolumn{3}{c}{\textbf{This-is-my (Single)}~\cite{seifi2025personalization}} &
\multicolumn{3}{c}{\textbf{This-is-my (Multi)}} &
\multicolumn{3}{c}{\textbf{RAP (Multi)}~\cite{hao2025rap}} \\
\cmidrule(lr){4-6} \cmidrule(lr){7-9} \cmidrule(lr){10-12} \cmidrule(lr){13-15} \cmidrule(lr){16-18}
& & &Prec. $\uparrow$ & Rec. $\uparrow$ & F1 $\uparrow$ & Prec. $\uparrow$ & Rec. $\uparrow$& F1 $\uparrow$& Prec. $\uparrow$& Rec. $\uparrow$& F1 $\uparrow$& Prec. $\uparrow$& Rec. $\uparrow$& F1 $\uparrow$& Prec. $\uparrow$& Rec. $\uparrow$& F1 $\uparrow$\\
\midrule
\multicolumn{18}{c}{\textbf{1 Reference View}} \\
\midrule
RAP~\cite{hao2025rap} & Intern  & 24hrs & 63.4 &\underline{98.2} & 77.0 & 47.8 & \underline{93.7} & 63.3 & 83.4 & \textbf{91.3} & \textbf{87.1} & \textbf{100.0} & 62.0 & 76.5 & 90.7 & \textbf{100.0} & \underline{95.1} \\
R2P~\cite{das2025training} & Intern  & 5.98s & 54.1 & 93.4 & 68.5 & 53.1 & 85.6 & 65.5 & 61.0 & 76.3 & 67.7 & -- & -- & -- & -- & -- & -- \\
Ego (Ours) & Intern & \underline{1.40s} & \underline{86.0} & 94.8 & \underline{90.2} & \underline{77.2} & 83.4 & 80.2 & 81.3 & \underline{77.0} & \underline{79.1} & 93.9 & 78.2 & \underline{88.6} & \textbf{100.0} & \underline{96.9} & \textbf{98.4} \\
& Qwen & \textbf{1.25}s & 76.9 & 90.3 & 83.1 & 65.9 & 90.2 & 76.2 & 73.3 & 67.6 & 70.3 & 83.4 & \textbf{87.3} & 85.3 & \underline{94.1} & 78.1 & 85.4 \\
\midrule
\multicolumn{18}{c}{\textbf{5 Reference Views}} \\
\midrule
PeKit~\cite{seifi2025personalization} & -- & 21.3s & 82.3 & 97.6 & 89.2 & 74.8 & 91.0 & 82.1 & \textbf{90.1} & 69.0 & 78.1 & \underline{96.1} & 45.4 & 61.6 & -- & -- & -- \\
Ego (Ours) & Intern  & 7.00s & \textbf{87.7} & \textbf{99.0} & \textbf{92.8} & \textbf{85.0} & 86.4 & \textbf{85.7} & \underline{87.2} & 65.9 & 75.1 & \textbf{100.0} & \underline{81.8} & \textbf{90.9} & -- & -- & -- \\
& Qwen & 6.25s & 72.1 & 95.9 & 82.3 & 67.9 & \textbf{98.7} & 80.5 & 77.6 & 71.9 & 74.6 & 92.1 & 74.5 & 82.4 & -- & -- & -- \\
\bottomrule
\end{tabular}
}

\label{tab:recognition}
\end{table*}

\myparagraph{Personalization Tasks and Metrics}
We cover three tasks:
\begin{itemize}[leftmargin=*,nosep]
\item \textbf{Recognition.} 
Determine whether a concept is present in the query image, responding with \textit{Yes} or \textit{No}. ecognition metrics are computed using the ground-truth validation images of each concept as positive examples, while images from all other categories serve as negative examples for that concept. Each image is evaluated for every concept, assessing robustness to intra-category variation. We avoid pooled negatives~\cite{alaluf2024myvlm,nguyen2024yo} for reproducibility.

\textbf{Metrics:} Previous works have adopted different metrics for evaluating recognition accuracy, including positive accuracy (Recall), negative accuracy (Specificity), and their weighted combination~\cite{alaluf2024myvlm, nguyen2024yo}. PeKit~\cite{seifi2025personalization} additionally incorporates Precision into its reporting. In line with ~\cite{hao2025rap}, we standardize all comparisons using three conventional binary classification metrics: \texttt{Precision}, \texttt{Recall}, and \texttt{F1-score}. For a single concept, \texttt{Precision} represents the fraction of true positives among all predicted positives $\frac{\text{TP}}{\text{TP} + \text{FP}}.$
{Recall} captures the fraction of true positives among all ground-truth positives $ \frac{\text{TP}}{\text{TP} + \text{FN}}.$
The {F1-score} is the harmonic mean of Precision and Recall
$ 2 \times \frac{\text{Precision} \times \text{Recall}}{\text{Precision} + \text{Recall}}.$

Overall performance is reported by averaging Precision and Recall across all concepts in the dataset, and computing the F1-score from these averaged values. For multi-concept, we compute the same metrics for each concept pair, where a model's response is considered correct only if it identifies all concepts present in the query image.

\item \textbf{Visual Question Answering (VQA).} The model is presented with a question concerning a specific concept(s) within the query image, expressed through their personalized names. The question may be formulated as either multiple-choice or open-ended. This task assesses the model’s ability to effectively collaborate with the user by accurately interpreting and reasoning on personalized references. \textbf{Metrics:} We measure \texttt{accuracy} as the fraction of questions answered correctly. For multiple-choice questions, we determine correctness through  string matching with the ground-truth answer. For open-ended questions, we use ChatGPT to automatically assess whether the model’s response is semantically and contextually aligned with the ground-truth answer, following an evaluation protocol similar to~\cite{maaz2024video}.
\item \textbf{Captioning.}  the model is required to generate a textual description of the query image that correctly identifies and incorporates the relevant concept name from all available concepts in the dataset. Unlike the recognition and VQA tasks, where the query is constructed based on a known ground-truth concept for a given query and no retrieval is necessary, this task additionally evaluates the model’s ability to select the appropriate concept for a given image before generating the caption. \textbf{Metrics:} We employ \texttt{captioning recall}, calculated as the fraction of query images for which the generated caption correctly references the concept or concept-pair shown in the image, with results averaged across all concepts.
\end{itemize}


\myparagraph{Datasets}.
We cover a diverse suite of datasets designed to assess the methods’  adaptability across \textit{single-concept}, \textit{multi-concept}, and \textit{video-based} personalization scenarios.
\begin{itemize}[leftmargin=*,nosep]
\item \textbf{Single-concept. } We utilize three datasets: \textit{MyVLM}~\cite{alaluf2024myvlm}, \textit{Yo'LLaVA}~\cite{nguyen2024yo}, and \textit{This-is-my-img}~\cite{seifi2025personalization}. The \textit{MyVLM} dataset comprises 29 object categories, and we evaluate the single-concept captioning task on this dataset. \textit{Yo'LLaVA} offers broader coverage, featuring 40 categories including objects, cartoon characters, Vietnamese public figures, and architectural landmarks; it also provides  multiple-choice (A/B) questions for the VQA task. The \textit{This-is-my-img} dataset presents 14 object and person categories collected from YouTube videos, emphasizing personalized visual understanding under in-the-wild conditions, in addition to  single-concept multiple-choice (A/B) VQA questions.
To ensure consistent evaluation across all methods, we establish fixed training (reference views) splits with $1$ and $5$  images per dataset. 
\item \textbf{Multi-concept. } We use both the This-is-my-img~\cite{seifi2025personalization} and RAP~\cite{hao2025rap} datasets. The multi-concept extension of This-is-my-img expands the original dataset by adding co-occurring people and objects, covering 22 concepts within 11 concept pairs, this is accompanied by a question–answer pair for  open-ended VQA task. We evaluate the multi-concept VQA and captioning tasks on this split. The RAP dataset provides a multi-concept validation split spanning 16 concepts and 8 concept pairs, including people, cartoon characters, and animals, offering a challenging benchmark for multi-entity personalization.
\item  \textbf{Video.} We adopt the video question-answering  dataset introduced by~\cite{seifi2025personalization}. This evaluation is conducted on 267 question–answer pairs derived from the original validation segments of the \textit{This-is-my} dataset~\cite{yeh2023meta}, enabling analysis of the model’s temporal reasoning and consistency in personalized Video QA.
\end{itemize}

\myparagraph{Models}
Early methods for LVLM personalization were based on earlier, less performant LVLMs such as MiniGPT-4~\citep{zhu2023minigpt} and LLaVA-1.5~\citep{liu2024improved}. Such models could not process multiple images and lacked visual in-context learning capability~\citep{kang2025online}. In this evaluation, we focus on more powerful LVLMs that demonstrate strong capabilities across a wide range of vision tasks~\footnote{VLM ranking: \href{https://mmbench.opencompass.org.cn/leaderboard}{MMBench}.}. Such LVLMs inherently possess better generalization and reasoning capabilities, making effective personalization especially valuable for real-world and complex visual understanding tasks. We evaluate SOTA personalization methods when paired with stronger, stable models while maintaining a feasible scale of model parameters for practical use cases. Namely, we adopt \textbf{InternVL3-14B}~\cite{zhu2025internvl3} as the primary LVLM for our method and for all reproduced baselines.
We further evaluate our method when paired with a smaller model, \textbf{Qwen2.5-VL-7B-Instruct}~\cite{bai2025qwen2}, analyzing the relationship between model capacity and training-free personalization performance.\\
\myparagraph{Implementation Details.}
 We refer to the appendix for the extended implementation details.


\subsection{Recognition}
We evaluate recognition performance under two reference-view settings: (a) \textbf{1-view}, supported by RAP, R2P, and \ours, and (b) \textbf{5-views}, supported only by PeKit and \ours. As shown in Tab. \ref{tab:recognition}, \ours achieves the highest F1-scores across most datasets and settings, including challenging multi-concept scenarios that reflect real-world personalization. 
Moreover, \ours requires minimal concept processing time, relying only on brief descriptions rather than extensive reasoning or attribute selection.
\begin{table*}[ht!]
\centering
\vspace{-0.1cm}
\caption{\textbf{VQA Acc. and Captioning Recall.}
Performance comparison ( 1 ref. view and InternVL3). \ours\ attains competitive accuracy on single-concept VQA while achieving SOTA performance in both multi-concept and video settings. \ours delivers superior captioning-recall. R2P only supports single-concept personalization and RAP does not support video personalization.}
\resizebox{0.9\textwidth}{!}{
\begin{tabular}{lc
>{\columncolor{singleconcept1}}c
>{\columncolor{singleconcept2}}c
>{\columncolor{singleconcept3}}c
>{\columncolor{multiconcept1}}c
>{\columncolor{multiconcept2}}c
>{\columncolor{videocolor}}c
}
\toprule
& \multicolumn{1}{c}{} &
\multicolumn{3}{c}{\textbf{Single Concept}} &
\multicolumn{2}{c}{\textbf{Multi Concept}} &
\multicolumn{1}{c}{\textbf{Video}} \\
\cmidrule(lr){3-5} \cmidrule(lr){6-7} \cmidrule(lr){8-8}
\textbf{Method} &
\makecell{\textbf{Sample} \\ \textbf{Runtime (s $\downarrow$)}} &
\multicolumn{2}{>{\columncolor{singleconcept1}}c}{\textbf{VQA (Acc. $\uparrow$)}} &
\textbf{Captioning (Recall $\uparrow$)} &
\textbf{VQA (Acc. $\uparrow$)} &
\textbf{Captioning (Recall $\uparrow$)} &
\textbf{VQA (Acc. $\uparrow$)} \\
\cmidrule(lr){3-4} \cmidrule(lr){5-5} \cmidrule(lr){6-6} \cmidrule(lr){7-7} \cmidrule(lr){8-8}
& &
\textbf{Yo’LLaVA} &
\textbf{This-is-my} &
\textbf{MyVLM} &
\textbf{This-is-my} &
\textbf{This-is-my} &
\textbf{This-is-my}\\
\midrule
RAP~\cite{hao2025rap}       & 7.8 & \textbf{97.6}       & \underline{90.0}      & 65.6              & \underline{53.7}  & \underline{43.6}  & -- \\
R2P~\cite{das2025training}  & 7.0 & 94.0                & 82.0                  & 77.5              & --                & --                & -- \\
PeKit              & \textbf{5.8} & \underline{94.6}    & \textbf{92.0}         & \underline{81.1}  & 51.8              & 35.2              & \underline{59.9} \\
\ours (Ours)      & \underline{6.0} & 92.3                & 88.0                  & \textbf{91.3}     & \textbf{72.2}     & \textbf{70.9}     & \textbf{70.0} \\
\bottomrule
\end{tabular}
}

\label{tab:vqa_caption}
\vspace{-0.3cm}
\end{table*}
RAP, despite large-scale finetuning (210k samples) and full reference views, is constrained by top-$k$ retrieval (set to 3), limiting detectable concepts and causing in-context saturation. While RAP excels in Recall, its low Precision reveals a tendency to over-predict, a byproduct of finetuning. Performance drops sharply in multi-concept cases, since the training dataset doesn't represent this case, underscoring the dependency of training-based methods on curated datasets.
R2P, though training-free, inherits the same top-$k$ restriction and fails in multi-concept settings with notably low precision.
PeKit performs strongly in single-concept detection but struggles in multi-concept cases because it relies on a \emph{single fixed similarity threshold}, which  introduces imbalance: a value suitable for one concept may be overly strict or lenient for another, leading to false negatives and significantly reducing recall, despite maintaining high precision in multi-concept scenarios.
\textit{Ego} overcomes these limitations by extracting compact, discriminative visual memory from the VLM’s embedding space, guided by its internal attention scores. This directs the model toward the most informative regions from the \textbf{model's own view}, filtering out background clutter and avoiding the distractions introduced by full reference images, a common failure mode in RAP and R2P. On RAP, \textit{Ego} improves F1-score by $3.3\%$, and on the more challenging This-is-my dataset the margin increases to \textbf{12\%}, showing robustness to occlusion, blur, and other real-world factors. Its ability to leverage multiple reference views further strengthens performance in complex multi-concept settings.

In our protocol, precision and recall are computed over all dataset examples, where positives represent only a small fraction—roughly 1/\text{\#concepts}. As a result, predicting a non-existent concept severely penalizes precision, as reflected in baseline performance. This design mirrors real-world conditions, where personalized concepts are rare across diverse inputs. A robust model should therefore predict concepts accurately and default to generic outputs under uncertainty. \ours achieves a strong precision–recall balance and demonstrates scalability by delivering competitive results even with the smaller Qwen2.5VL backbone. Full comparisons to base models are provided in the Appendix.

\begin{figure*}[htbp]
      \centering
      \includegraphics[width=0.9\textwidth]{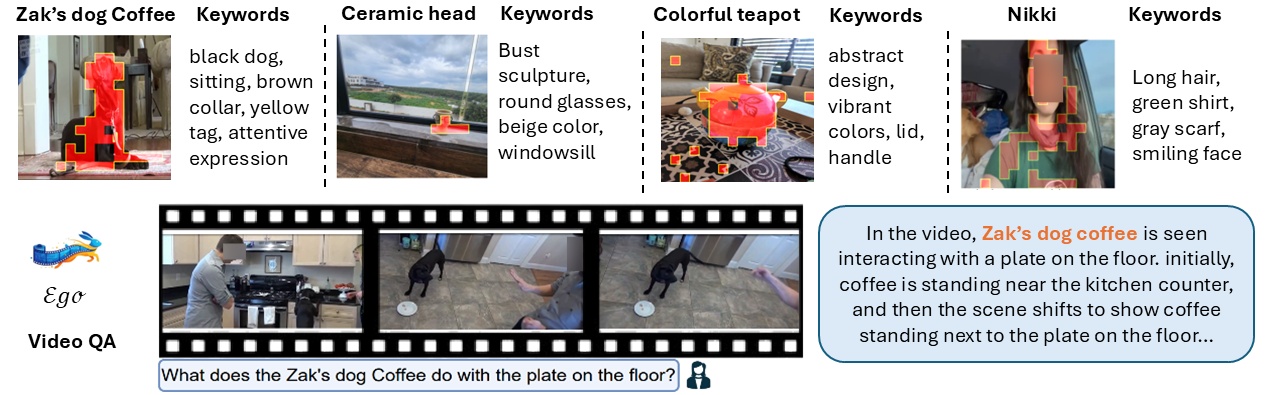}
      \vspace{-0.2cm}
      \caption{
\textbf{Qualitative results.} Top row: keywords and highlighted patches of selected visual tokens (\ours concept memory) for various concepts, illustrating their representativeness and adaptability to object size. Bottom row: \ours demonstrates Video QA capability. }
\label{fig:qualitative}

\end{figure*}
\subsection{Visual Question Answering}
Tab.~\ref{tab:vqa_caption} reports VQA performance in both single- and multi-concept settings. In the single-concept regime, \ours performs close to RAP, even though RAP is trained on data containing VQA-specific supervision. Compared to R2P, \ours underperforms on Yo’LLaVA but achieves a clear improvement on the more challenging This-is-my dataset.

The advantages of \ours become more evident in the multi-concept scenario. On the multi-concept split of This-is-my, \ours exceeds RAP by nearly 20\%, reflecting its ability to preserve multiple personalized concepts. Moreover, \ours applies directly to video personalization without any modifications and outperforms the PeKit pipeline, underscoring its generality across personalization modalities.

\subsection{Captioning Recall}

In Tab.~\ref{tab:vqa_caption} we present our results on personalized captioning. We observe a clear advantage of \ours over prior methods. In  MyVLM dataset, \ours improves performance by 14\% over R2P, while in the challenging multi-concept setting of This-is-my, it achieves nearly 30\% improvement over RAP. Notably, \ours\ also achieves faster inference by using a precomputed, informative subset of visual tokens from the reference views for in-context conditioning, avoiding the need to process full reference images with the vision encoder. Unlike previous methods that rely on full reference views and external retrieval modules, \ours leverages the VLM’s inherent capabilities: (1) strong discriminative power to extract unique visual features for each concept while filtering out background noise, and (2) in-context learning to build an In-Context Memory of concepts and retrieve the correct concept internally through attention during caption generation. This parallel, integrated approach reduces latency, eliminates additional stages, and improves retrieval quality—highlighting the benefit of our method over using the raw reference view, an advantage that was not apparent in simpler tasks such as recognition and VQA.


\subsection{Ablation}
We analyze the main design choices of \ours. 
Our method relies on keywords produced by the model to extract key visual tokens representing the model's memory of a concept. We evaluate the performance of the model provided with the keywords as the concept memory instead (\textbf {Keywords Only}). 
To evaluate the quality of our visual token selection, we consider two baselines, uniform selection of $K$ visual tokens (Sec.\ref{sec:memsize}) from the reference view (\textbf{Uniform Visual)}, and using all the visual tokens in the reference view (\textbf{Full Visual}). $K$ here is set to $20\%$ for both Uniform and \ours. We also consider the combination of keywords and the full visual tokens (\textbf{Full Visual + Keywords}).

Results (F1 score) based on  InternVL3 (14B) and the Yo'LLaVA dataset in the  recognition task are reported in Tab.~\ref{tab:ablation}.
It can be observed that descriptive words alone lack the discriminative power to strongly identify the concept, compared to the visual tokens counterpart, and it does not provide any additional useful information but rather a distraction when combined with full visual tokens of the concept. This confirms the importance of constructing a \textbf{visual  concept memory}. \ours significantly outperforms Uniform. Notably, \ours with 5 reference views improves significantly over the Full Visual~(+1.7\%) while using the same number of visual tokens.  This indicates that our attention-based selection strategy succeeds in identifying the key visual patches of a concept.
We refer to the following analysis and ablations in the Appendix: 1) The effect of  dynamic concept memory size $K_c$ versus the default size $K$~(Sec.\ref{sec:memsize}). 2) Using cross attention to full concept description rather than the keywords, showing the utility of keywords selection. 3) Repeated sampling from the LLM output during keywords generation with results showing no significant impact. 4) Our automatic layer selection strategy~(Sec.\ref{sec:layerselction}) compared to tuned layer selection based on downstream performance, showing the stability of our layer selection on the two models. 5) Comparison to the LVLM with full reference views provided in-context on various tasks, showing the effectiveness and efficiency of \ours. Note that this baseline does not scale to cases with many personalization concepts. 6) An evaluation of \ours across smaller‑scale LVLMs to assess how model size influences personalization performance.
7) A comparison between our attention‑guided embedding extraction and an alternative approach that uses full segmentation masks, highlighting the benefits of our proposed method.



\begin{table}[ht]
\centering
\caption{\textbf{Ablation}.  \ours outperforms uniform visual token selection and Full Visual under the same In Context tokens budget.}
\vspace{-0.2cm}
\resizebox{\columnwidth}{!}{
\begin{tabular}{l|cc|c}
\toprule
\textbf{Method} & \textbf{\% of Visual Tokens} & \textbf{Word Injection} & \textbf{F1 Score $\uparrow$} \\
\midrule
Keywords    & \XSolidBrush & \Checkmark   & 71.3 \\
Full Visual             & 100\%        & \XSolidBrush & \underline{84.1} \\
Full Visual + Keywords      & 100\%        & \Checkmark   & 82.5\\
Uniform  & 20\%         & \XSolidBrush   & 77.7 \\
\hline
\ours (1-view)     & 20\%         & \XSolidBrush   & 80.4 \\
\hline
\ours (5-view)     & 20\%         & \XSolidBrush   & \textbf{85.7} \\
\bottomrule
\end{tabular}
}
\label{tab:ablation}
\end{table}

\subsection{Qualitative Results}
Fig.~\ref{fig:qualitative} illustrates how our attention mechanism localizes representative visual tokens in reference views from the This-is-my~\cite{yeh2023meta} and MyVLM~\cite{alaluf2024myvlm} datasets. It can be seen how our dynamic size selection strategy effectively reduces redundant tokens and background noise. It also illustrates \ours ability to track concepts in a video and answer questions accordingly. Further qualitative results are in the Appendix.

\section{Discussion}
In this work, we aim to establish a strong and efficient post-hoc LVLM personalization method. Our approach assumes that the LVLM has robust visual understanding capabilities and may not perform well with older models. However, given the availability of powerful open-source alternatives, we do not anticipate the need to rely on less performant models. It is worth noting that full reliance on the LVLM requires an instruction prompt tailored to the specific model. 

For evaluation, we prioritized fairness and reproducibility: we re-evaluated major personalization baselines on the same datasets under a unified set of expressive metrics. Our experiments cover most existing personalization datasets across diverse tasks—Recognition, VQA, and Captioning—as well as different personalization scenarios, including single-concept, multi-concept, and video.

\ours demonstrates a strong balance between efficiency and performance, outperforming state-of-the-art methods and baselines. Nevertheless, our results indicate significant room for improvement. We envision our evaluation protocol serving as a testbed for future personalization research.

{
    \small
    \bibliographystyle{ieeenat_fullname}
    \bibliography{main}

\begin{thebibliography}{39}
\providecommand{\natexlab}[1]{#1}
\providecommand{\url}[1]{\texttt{#1}}
\expandafter\ifx\csname urlstyle\endcsname\relax
  \providecommand{\doi}[1]{doi: #1}\else
  \providecommand{\doi}{doi: \begingroup \urlstyle{rm}\Url}\fi

\bibitem[Alaluf et~al.(2024)Alaluf, Richardson, Tulyakov, Aberman, and Cohen-Or]{alaluf2024myvlm}
Yuval Alaluf, Elad Richardson, Sergey Tulyakov, Kfir Aberman, and Daniel Cohen-Or.
\newblock Myvlm: Personalizing vlms for user-specific queries.
\newblock In \emph{European Conference on Computer Vision}, pages 73--91. Springer, 2024.

\bibitem[Bai et~al.(2025)Bai, Chen, Liu, Wang, Ge, Song, Dang, Wang, Wang, Tang, et~al.]{bai2025qwen2}
Shuai Bai, Keqin Chen, Xuejing Liu, Jialin Wang, Wenbin Ge, Sibo Song, Kai Dang, Peng Wang, Shijie Wang, Jun Tang, et~al.
\newblock Qwen2. 5-vl technical report.
\newblock \emph{arXiv preprint arXiv:2502.13923}, 2025.

\bibitem[Baldassini et~al.(2024)Baldassini, Shukor, Cord, Soulier, and Piwowarski]{baldassini2024makes}
Folco~Bertini Baldassini, Mustafa Shukor, Matthieu Cord, Laure Soulier, and Benjamin Piwowarski.
\newblock What makes multimodal in-context learning work?
\newblock In \emph{Proceedings of the IEEE/CVF Conference on Computer Vision and Pattern Recognition}, pages 1539--1550, 2024.

\bibitem[Chen et~al.(2024)Chen, Zhao, Liu, Bai, Lin, Zhou, and Chang]{chen2024image}
Liang Chen, Haozhe Zhao, Tianyu Liu, Shuai Bai, Junyang Lin, Chang Zhou, and Baobao Chang.
\newblock An image is worth 1/2 tokens after layer 2: Plug-and-play inference acceleration for large vision-language models.
\newblock In \emph{European Conference on Computer Vision}, pages 19--35. Springer, 2024.

\bibitem[Chen et~al.(2025)Chen, Liu, Han, Xia, Cremers, Torr, Tresp, and Gu]{chen2025true}
Shuo Chen, Jianzhe Liu, Zhen Han, Yan Xia, Daniel Cremers, Philip Torr, Volker Tresp, and Jindong Gu.
\newblock True multimodal in-context learning needs attention to the visual context.
\newblock In \emph{Second Conference on Language Modeling}, 2025.

\bibitem[Cohen et~al.(2022)Cohen, Gal, Meirom, Chechik, and Atzmon]{cohen2022my}
Niv Cohen, Rinon Gal, Eli~A Meirom, Gal Chechik, and Yuval Atzmon.
\newblock “this is my unicorn, fluffy”: Personalizing frozen vision-language representations.
\newblock In \emph{European conference on computer vision}, pages 558--577. Springer, 2022.

\bibitem[Das et~al.(2025)Das, Talon, Wang, Mancini, and Ricci]{das2025training}
Deepayan Das, Davide Talon, Yiming Wang, Massimiliano Mancini, and Elisa Ricci.
\newblock Training-free personalization via retrieval and reasoning on fingerprints.
\newblock \emph{arXiv preprint arXiv:2503.18623}, 2025.

\bibitem[Dorovatas et~al.()Dorovatas, Seifi, Gupta, and Aljundi]{dorovatasrecurrent}
Vaggelis Dorovatas, Soroush Seifi, Gunshi Gupta, and Rahaf Aljundi.
\newblock Recurrent attention-based token selection for efficient streaming video-llms.
\newblock In \emph{The Thirty-ninth Annual Conference on Neural Information Processing Systems}.

\bibitem[Fan et~al.(2025)Fan, Zhao, Fu, Tong, Su, Pan, Zhang, and Shen]{fan-etal-2025-visipruner}
Yingqi Fan, Anhao Zhao, Jinlan Fu, Junlong Tong, Hui Su, Yijie Pan, Wei Zhang, and Xiaoyu Shen.
\newblock {V}isi{P}runer: Decoding discontinuous cross-modal dynamics for efficient multimodal {LLM}s.
\newblock In \emph{Proceedings of the 2025 Conference on Empirical Methods in Natural Language Processing}, pages 18896--18913, Suzhou, China, 2025. Association for Computational Linguistics.

\bibitem[Gal et~al.(2022)Gal, Alaluf, Atzmon, Patashnik, Bermano, Chechik, and Cohen-Or]{gal2022image}
Rinon Gal, Yuval Alaluf, Yuval Atzmon, Or Patashnik, Amit~H Bermano, Gal Chechik, and Daniel Cohen-Or.
\newblock An image is worth one word: Personalizing text-to-image generation using textual inversion.
\newblock \emph{arXiv preprint arXiv:2208.01618}, 2022.

\bibitem[Hao et~al.(2025)Hao, Han, Li, Li, and Yue]{hao2025rap}
Haoran Hao, Jiaming Han, Changsheng Li, Yu-Feng Li, and Xiangyu Yue.
\newblock Rap: Retrieval-augmented personalization for multimodal large language models.
\newblock In \emph{Proceedings of the Computer Vision and Pattern Recognition Conference}, pages 14538--14548, 2025.

\bibitem[Huang et~al.(2025)Huang, Zhou, and Han]{huang2025llmvtp}
Xiaohu Huang, Hao Zhou, and Kai Han.
\newblock {LLM}-{VTP}: {LLM}-reasoned visual token pruning for efficient multi-modal video understanding, 2025.

\bibitem[Jiang et~al.(2025)Jiang, Chen, Zhu, Luo, Shen, and Yang]{jiang2025devils}
Zhangqi Jiang, Junkai Chen, Beier Zhu, Tingjin Luo, Yankun Shen, and Xu Yang.
\newblock Devils in middle layers of large vision-language models: Interpreting, detecting and mitigating object hallucinations via attention lens.
\newblock In \emph{Proceedings of the Computer Vision and Pattern Recognition Conference}, pages 25004--25014, 2025.

\bibitem[Johnson et~al.(2019)Johnson, Douze, and J{\'e}gou]{johnson2019billion}
Jeff Johnson, Matthijs Douze, and Herv{\'e} J{\'e}gou.
\newblock Billion-scale similarity search with gpus.
\newblock \emph{IEEE Transactions on Big Data}, 7\penalty0 (3):\penalty0 535--547, 2019.

\bibitem[Kang et~al.(2025)Kang, Aljundi, Dorovatas, and Alahari]{kang2025online}
Zhiqi Kang, Rahaf Aljundi, Vaggelis Dorovatas, and Karteek Alahari.
\newblock Online in-context distillation for low-resource vision language models.
\newblock \emph{arXiv preprint arXiv:2510.18117}, 2025.

\bibitem[Kim et~al.(2025)Kim, Kang, Park, Kim, and Hwang]{kim2025interpreting}
Jinyeong Kim, Seil Kang, Jiwoo Park, Junhyeok Kim, and Seong~Jae Hwang.
\newblock Interpreting attention heads for image-to-text information flow in large vision-language models.
\newblock \emph{arXiv preprint arXiv:2509.17588}, 2025.

\bibitem[Lai et~al.(2024)Lai, Gan, Wu, Qi, and Yu]{lai2024large}
Jinqi Lai, Wensheng Gan, Jiayang Wu, Zhenlian Qi, and Philip~S Yu.
\newblock Large language models in law: A survey.
\newblock \emph{AI Open}, 5:\penalty0 181--196, 2024.

\bibitem[Li et~al.(2023)Li, Li, Savarese, and Hoi]{li2023blip}
Junnan Li, Dongxu Li, Silvio Savarese, and Steven Hoi.
\newblock Blip-2: Bootstrapping language-image pre-training with frozen image encoders and large language models.
\newblock In \emph{International conference on machine learning}, pages 19730--19742. PMLR, 2023.

\bibitem[Lin et~al.(2014)Lin, Maire, Belongie, Hays, Perona, Ramanan, Doll{\'a}r, and Zitnick]{lin2014microsoft}
Tsung-Yi Lin, Michael Maire, Serge Belongie, James Hays, Pietro Perona, Deva Ramanan, Piotr Doll{\'a}r, and C~Lawrence Zitnick.
\newblock Microsoft coco: Common objects in context.
\newblock In \emph{European Conference on Computer Vision (ECCV)}, pages 740--755, 2014.

\bibitem[Liu et~al.(2023)Liu, Li, Wu, and Lee]{liu2023visual}
Haotian Liu, Chunyuan Li, Qingyang Wu, and Yong~Jae Lee.
\newblock Visual instruction tuning.
\newblock \emph{Advances in neural information processing systems}, 36:\penalty0 34892--34916, 2023.

\bibitem[Liu et~al.(2024)Liu, Li, Li, and Lee]{liu2024improved}
Haotian Liu, Chunyuan Li, Yuheng Li, and Yong~Jae Lee.
\newblock Improved baselines with visual instruction tuning.
\newblock In \emph{Proceedings of the IEEE/CVF conference on computer vision and pattern recognition}, pages 26296--26306, 2024.

\bibitem[Maaz et~al.(2024)Maaz, Rasheed, Khan, and Khan]{maaz2024video}
Muhammad Maaz, Hanoona Rasheed, Salman Khan, and Fahad Khan.
\newblock Video-chatgpt: Towards detailed video understanding via large vision and language models.
\newblock In \emph{Proceedings of the 62nd Annual Meeting of the Association for Computational Linguistics (Volume 1: Long Papers)}, pages 12585--12602, 2024.

\bibitem[Mann et~al.(2020)Mann, Ryder, Subbiah, Kaplan, Dhariwal, Neelakantan, Shyam, Sastry, Askell, Agarwal, et~al.]{mann2020language}
Ben Mann, Nick Ryder, Melanie Subbiah, J Kaplan, P Dhariwal, A Neelakantan, P Shyam, G Sastry, A Askell, S Agarwal, et~al.
\newblock Language models are few-shot learners.
\newblock \emph{arXiv preprint arXiv:2005.14165}, 1\penalty0 (3):\penalty0 3, 2020.

\bibitem[Nguyen et~al.(2024)Nguyen, Liu, Li, Cai, Ojha, and Lee]{nguyen2024yo}
Thao Nguyen, Haotian Liu, Yuheng Li, Mu Cai, Utkarsh Ojha, and Yong~Jae Lee.
\newblock Yo'llava: Your personalized language and vision assistant.
\newblock \emph{Advances in Neural Information Processing Systems}, 37:\penalty0 40913--40951, 2024.

\bibitem[Pham et~al.(2024)Pham, Phan, Doermann, and Tian]{pham2024personalized}
Chau Pham, Hoang Phan, David Doermann, and Yunjie Tian.
\newblock Personalized large vision-language models.
\newblock \emph{arXiv preprint arXiv:2412.17610}, 2024.

\bibitem[Pi et~al.(2024)Pi, Zhang, Han, Zhang, Pan, and Zhang]{pi2024personalized}
Renjie Pi, Jianshu Zhang, Tianyang Han, Jipeng Zhang, Rui Pan, and Tong Zhang.
\newblock Personalized visual instruction tuning.
\newblock \emph{arXiv preprint arXiv:2410.07113}, 2024.

\bibitem[Radford et~al.(2021)Radford, Kim, Hallacy, Ramesh, Goh, Agarwal, Sastry, Askell, Mishkin, Clark, et~al.]{radford2021learning}
Alec Radford, Jong~Wook Kim, Chris Hallacy, Aditya Ramesh, Gabriel Goh, Sandhini Agarwal, Girish Sastry, Amanda Askell, Pamela Mishkin, Jack Clark, et~al.
\newblock Learning transferable visual models from natural language supervision.
\newblock In \emph{International conference on machine learning}, pages 8748--8763. PmLR, 2021.

\bibitem[Ren et~al.(2024)Ren, Liu, Zeng, Lin, Li, Cao, Chen, Huang, Chen, Yan, et~al.]{ren2024grounded}
Tianhe Ren, Shilong Liu, Ailing Zeng, Jing Lin, Kunchang Li, He Cao, Jiayu Chen, Xinyu Huang, Yukang Chen, Feng Yan, et~al.
\newblock Grounded sam: Assembling open-world models for diverse visual tasks.
\newblock \emph{arXiv preprint arXiv:2401.14159}, 2024.

\bibitem[Ruiz et~al.(2023)Ruiz, Li, Jampani, Pritch, Rubinstein, and Aberman]{ruiz2023dreambooth}
Nataniel Ruiz, Yuanzhen Li, Varun Jampani, Yael Pritch, Michael Rubinstein, and Kfir Aberman.
\newblock Dreambooth: Fine tuning text-to-image diffusion models for subject-driven generation.
\newblock In \emph{Proceedings of the IEEE/CVF conference on computer vision and pattern recognition}, pages 22500--22510, 2023.

\bibitem[Ruiz et~al.(2024)Ruiz, Li, Jampani, Wei, Hou, Pritch, Wadhwa, Rubinstein, and Aberman]{ruiz2024hyperdreambooth}
Nataniel Ruiz, Yuanzhen Li, Varun Jampani, Wei Wei, Tingbo Hou, Yael Pritch, Neal Wadhwa, Michael Rubinstein, and Kfir Aberman.
\newblock Hyperdreambooth: Hypernetworks for fast personalization of text-to-image models.
\newblock In \emph{Proceedings of the IEEE/CVF conference on computer vision and pattern recognition}, pages 6527--6536, 2024.

\bibitem[Schick et~al.(2023)Schick, Dwivedi-Yu, Dess{\`\i}, Raileanu, Lomeli, Hambro, Zettlemoyer, Cancedda, and Scialom]{schick2023toolformer}
Timo Schick, Jane Dwivedi-Yu, Roberto Dess{\`\i}, Roberta Raileanu, Maria Lomeli, Eric Hambro, Luke Zettlemoyer, Nicola Cancedda, and Thomas Scialom.
\newblock Toolformer: Language models can teach themselves to use tools.
\newblock \emph{Advances in Neural Information Processing Systems}, 36:\penalty0 68539--68551, 2023.

\bibitem[Seifi et~al.(2025)Seifi, Dorovatas, Reino, and Aljundi]{seifi2025personalization}
Soroush Seifi, Vaggelis Dorovatas, Daniel~Olmeda Reino, and Rahaf Aljundi.
\newblock Personalization toolkit: Training free personalization of large vision language models.
\newblock \emph{arXiv preprint arXiv:2502.02452}, 2025.

\bibitem[Shah et~al.(2023)Shah, Osi{\'n}ski, Levine, et~al.]{shah2023lm}
Dhruv Shah, B{\l}a{\.z}ej Osi{\'n}ski, Sergey Levine, et~al.
\newblock Lm-nav: Robotic navigation with large pre-trained models of language, vision, and action.
\newblock In \emph{Conference on robot learning}, pages 492--504. PMLR, 2023.

\bibitem[Singhal et~al.(2023)Singhal, Azizi, Tu, Mahdavi, Wei, Chung, Scales, Tanwani, Cole-Lewis, Pfohl, et~al.]{singhal2023large}
Karan Singhal, Shekoofeh Azizi, Tao Tu, S~Sara Mahdavi, Jason Wei, Hyung~Won Chung, Nathan Scales, Ajay Tanwani, Heather Cole-Lewis, Stephen Pfohl, et~al.
\newblock Large language models encode clinical knowledge.
\newblock \emph{Nature}, 620\penalty0 (7972):\penalty0 172--180, 2023.

\bibitem[Wu et~al.(2024)Wu, Qiu, Zheng, Zhu, and Chen]{wu2024exploring}
Likang Wu, Zhaopeng Qiu, Zhi Zheng, Hengshu Zhu, and Enhong Chen.
\newblock Exploring large language model for graph data understanding in online job recommendations.
\newblock In \emph{Proceedings of the AAAI conference on artificial intelligence}, pages 9178--9186, 2024.

\bibitem[Yeh et~al.(2023)Yeh, Russell, Sivic, Heilbron, and Jenni]{yeh2023meta}
Chun-Hsiao Yeh, Bryan Russell, Josef Sivic, Fabian~Caba Heilbron, and Simon Jenni.
\newblock Meta-personalizing vision-language models to find named instances in video.
\newblock In \emph{Proceedings of the IEEE/CVF Conference on Computer Vision and Pattern Recognition}, pages 19123--19132, 2023.

\bibitem[Zhang et~al.(2024)Zhang, Rossi, Kveton, Shao, Yang, Zamani, Dernoncourt, Barrow, Yu, Kim, et~al.]{zhang2024personalization}
Zhehao Zhang, Ryan~A Rossi, Branislav Kveton, Yijia Shao, Diyi Yang, Hamed Zamani, Franck Dernoncourt, Joe Barrow, Tong Yu, Sungchul Kim, et~al.
\newblock Personalization of large language models: A survey.
\newblock \emph{arXiv preprint arXiv:2411.00027}, 2024.

\bibitem[Zhu et~al.(2023)Zhu, Chen, Shen, Li, and Elhoseiny]{zhu2023minigpt}
Deyao Zhu, Jun Chen, Xiaoqian Shen, Xiang Li, and Mohamed Elhoseiny.
\newblock Minigpt-4: Enhancing vision-language understanding with advanced large language models.
\newblock \emph{arXiv preprint arXiv:2304.10592}, 2023.

\bibitem[Zhu et~al.(2025)Zhu, Wang, Chen, Liu, Ye, Gu, Tian, Duan, Su, Shao, et~al.]{zhu2025internvl3}
Jinguo Zhu, Weiyun Wang, Zhe Chen, Zhaoyang Liu, Shenglong Ye, Lixin Gu, Hao Tian, Yuchen Duan, Weijie Su, Jie Shao, et~al.
\newblock Internvl3: Exploring advanced training and test-time recipes for open-source multimodal models.
\newblock \emph{arXiv preprint arXiv:2504.10479}, 2025.

\end{thebibliography}
}
\clearpage
\appendix
\appendixpage
\addappheadtotoc
\begin{appendices}
\section{Implementation Details}
All experiments are conducted at a fixed input resolution of $448 \times 448$ across all datasets and methods. Images are normalized using the LVLM’s default preprocessing pipeline, and tiling is disabled (set to 1) to minimize computational overhead. R2P~\cite{das2025training} and RAP~\cite{hao2025rap} are adapted to InternVL3-14B based on their official Github implementations, while PeKit~\cite{seifi2025personalization} is re-implemented from scratch following the original paper. Inference is performed with $2 \times$~A100 GPUs. Fine-tuning InternVL3-14B with RAP~\cite{hao2025rap} is carried out using $8 \times$~A100 GPUs, a batch size of 6, and a LoRA rank of~32. No gradient updates are applied for the remaining methods. For RAP~\cite{hao2025rap}, cosine similarity is used as the similarity metric, and a \texttt{top-k = 3} concept candidate retrieval is adopted for evaluation where applicable. All experiments are conducted using Pytorch 2.9. \ours's implementation will be available on Github.
\section{Ablation}
In this section, we present ablation studies to assess the impact of key design components in \ours. Most experiments in this section are conducted on the \textbf{This-Is-My} dataset (single concept) the \textbf{InternVL3-14B} model, focusing primarily on the \textbf{recognition task}.

\subsection{Dynamic $K_{c}$ vs Fixed $K$}\label{supp:size_selection}
\begin{table}[ht]
\centering
\caption{$K$ vs $K_c$: \ours adapts to the size of the target concept in the reference views and boosts the performance compared to a fixed memory size selection strategy.}
\begin{tabular*}{\columnwidth}{@{\extracolsep{\fill}}lccc}
    \toprule
    \textbf{Experiment} & \textbf{Prec.} & \textbf{Rec.} & \textbf{F1} \\
    \midrule
    & \multicolumn{3}{c}{\textbf{All concepts}} \\
    \cmidrule(lr){2-4}
    $K_c \le  50$   & 81.3 & 77.0 & \textbf{79.1} \\
    $K = 50$     & 82.4 & 74.3 & 78.1 \\
    \midrule
    & \multicolumn{3}{c}{\textbf{Zak’s Dog Coffee}~\ref{fig:zak_dog_coffee_extraction}} \\
    \cmidrule(lr){2-4}
    $K_c \le  50$  & 81.8 & 34.6 & \textbf{48.6} \\
    $K = 50$     & 100.0 & 3.85 & 7.41 \\
    \bottomrule
\end{tabular*}



\label{tab:kc_vs_k}
\end{table}
In this section, we analyze the impact of using a dynamic concept memory size $K_c$---as introduced in Section~3.3 of the main paper---compared to enforcing a fixed memory size $K$ on the recognition task. 
We consider a baseline with a constant $K = 50$ for all concepts, represented using a single reference view. Note that \ours caps the number of visual tokens to the same $K=50$ to maintain a minimal number of visual tokens representing an concept.

\begin{figure}[htbp]
    \centering
      \includegraphics[width=1\columnwidth]{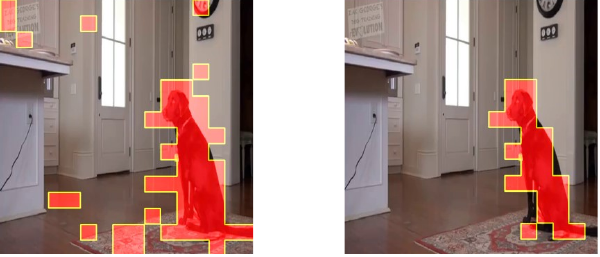}
      \caption{Extracted visual tokens for \textit{Zak's Dog Coffee}. \textbf{Left:} Fixed $K=50$ \textbf{Right}: Dynamic $K_c = 25$. \ours removes 25 background patches by adapting to the concept's size.}
    \label{fig:zak_dog_coffee_extraction}
\end{figure}

As shown in Table~\ref{tab:kc_vs_k}, the dynamic memory selection strategy outperforms the fixed-budget approach (while using less token budget). A representative example is the concept~\textit{Zak's Dog Coffee} from the This-is-My dataset.  As illustrated in Fig.~\ref{fig:zak_dog_coffee_extraction}, a fixed $K$ forces the model to store extra background tokens that may confuse the model for recognizing the object in novel environments. In contrast, the dynamic strategy enables \ours to effectively filter out such background noise, yielding higher recall and a substantially improved F1-score. This confirms that aligning the memory size with the visual footprint of the object is crucial for robust personalization, particularly for smaller objects.


\subsection{Layers Selection Strategy}
\begin{table}[ht]
\centering
\caption{LVLM Layer selection: Parentheses indicate the selected layers for InternVL3-14B. \ours outperforms the baselines by emphasizing more discriminative features while suppressing background noise.}
\begin{tabular*}{\columnwidth}{@{\extracolsep{\fill}}lccc}
    \toprule
    \textbf{Experiment} & \textbf{Prec.} & \textbf{Rec.} & \textbf{F1} \\
    \midrule
    \ours (29, 30, 35, 36, 39) & 81.3 & \textbf{77.0} & \textbf{79.1} \\
    Manual selection (20-24)    & \textbf{81.7} & 73.5 & 77.4 \\
    Uniform layer selection    & 78.7 & 73.1 & 75.8 \\
    \bottomrule
\end{tabular*}

\label{tab:layers}
\end{table}

In Section 3.4 of the main paper, we proposed an LVLM-agnostic strategy to select the layers that exhibit the strongest text–visual token interaction. We evaluate its effectiveness by comparing \ours with two baselines: manually choosing five consecutive intermediate layers inspired by prior work \cite{jiang2025devils}, and uniformly sampling five layers across the network.

As shown in Table \ref{tab:layers}, \ours surpasses the baselines, with uniform layer selection performing the worst. Manual selection exhibits a substantial decline in recall, highlighting the effectiveness of our layer selection strategy in achieving optimal performance.

It is important to note that personalized concepts typically occupy a large portion of the reference views in current personalization datasets. As a result, even a uniform selection of ~20\% of visual tokens is likely to include tokens overlapping with the target subject, resulting in a reasonable F1-score even when the selected LVLM layers are not tuned.

Nevertheless, the \textbf{3.3\%} gain achieved by \ours over the uniform layer selection represents the utility of our automatic layer selection  in identifying the key layers where visual tokens are interacting the most with the generated textual keywords.

\subsection{Full-caption vs Keywords Attention-based Selection}
As detailed in Sections 3.1 and 3.2 of the main paper, \ours leverages cross-modal attention maps to identify the most informative visual tokens that contribute to generating keywords describing the personalized concepts in the reference views. Table~\ref{tab:keywords} compares this approach with using full concept captions, which include generic terms (e.g., pronouns, articles) and sometimes background details. Although the performance gap is small in the 1-view setting, it becomes more pronounced with 5 views, indicating that full descriptions introduce cumulative noise from background tokens or attention sinks.
\begin{table}[ht]
\centering
\caption{Full description vs Keywords cross-attention on This-is-my single concept recognition task}
\begin{tabular*}{\columnwidth}{@{\extracolsep{\fill}}lccc}
    \toprule
    \textbf{Experiment} & \textbf{Prec.} & \textbf{Rec.} & \textbf{F1} \\
    \midrule
    & \multicolumn{3}{c}{\textbf{1 Ref. View}} \\
    \cmidrule(lr){2-4} 
    Keywords                 & \textbf{81.3} & 77.0 & \textbf{79.1}\\
    Caption & 80.4 & 77.6 & 79.0\\
    \midrule
    & \multicolumn{3}{c}{\textbf{5 Ref. views}} \\
    \cmidrule(lr){2-4}
    Keywords                 & 86.1 & \textbf{68.7} & \textbf{76.5} \\
    Caption & \textbf{87.2} & 65.9 & 75.1 \\
    \bottomrule
\end{tabular*}


\label{tab:keywords}
\end{table}

\subsection{Sampling for Keywords Attention-based Selection}
\begin{table}[ht]
\centering
\caption{Visual token selection over 5 samples vs no sampling on This-is-my single concept recognition task.}
\begin{tabular*}{\columnwidth}{@{\extracolsep{\fill}}lccc}
    \toprule
    \textbf{Experiment} & \textbf{Prec.} & \textbf{Rec.} & \textbf{F1} \\
    \midrule
    & \multicolumn{3}{c}{\textbf{1 Ref. View}} \\
    \cmidrule(lr){2-4}    
    No sampling & \textbf{81.3} & \textbf{77.0} & \textbf{79.1} \\
    $N=5$       & 81.2 & 75.9 & 78.5 \\
    \midrule
    & \multicolumn{3}{c}{\textbf{5 Ref. views}} \\
    \cmidrule(lr){2-4}
    No sampling & \textbf{87.2} & \textbf{65.9} & \textbf{75.1} \\
    $N=5$       & 86.8 & 63.5 & 73.4 \\
    \bottomrule
\end{tabular*}


\label{tab:sampling}
\end{table}

In this section, we assess whether increasing diversity in the descriptive keywords via sampling can improve the robustness of the selected visual memory, compared to a deterministic decoding strategy (temperature{=}0, do\_sample{=}False). We compare \ours against an aggregation-based baseline, in which keywords are sampled from five non-deterministic runs (do\_sample{=}True), and the resulting attention scores are averaged prior to selecting the top-$K_c$ visual tokens.

As shown in Table~\ref{tab:sampling}, the deterministic decoding strategy yields the strongest performance, suggesting that the most confident generated keywords describe the personalized concept better and are more reliably grounded in the visual content. Conversely, sampling introduces randomness that can produce weakly relevant or even hallucinated attributes. These noisy terms may divert attention toward irrelevant image regions, degrading the quality of the extracted concept memory and leading to reduced performance.


\subsection{\ours vs. Full reference Views and Base VLM}

\begin{table*}[h]
\centering
\caption{\textbf{VQA Accuracy and Captioning Recall.} Full Ref. View vs \ours vs Base VLM on all This-is-my tasks.}


\begin{tabular*}{\textwidth}{@{\extracolsep{\fill}}lcccccccccc}
\toprule
\multirow{2}{*}{\textbf{Method}} & 
\multirow{2}{*}{\textbf{Single VQA}} &
\multirow{2}{*}{\textbf{Multi VQA}} &
\multirow{2}{*}{\textbf{Video VQA}} &
\multicolumn{3}{c}{\textbf{Single Recognition}} &
\multicolumn{3}{c}{\textbf{Multi Recognition}} \\
\cmidrule(lr){5-7} \cmidrule(lr){8-10}
& & & & \textbf{Prec.} & \textbf{Rec.} & \textbf{F1} & \textbf{Prec.} & \textbf{Rec.} & \textbf{F1} \\
\midrule
\ours (Ours)    & \textbf{88.0} & \textbf{72.2} & \textbf{70.0} & 81.3 & \textbf{77.0} & \textbf{79.1} & 93.9 & \textbf{78.2} & \textbf{88.6} \\
Full Ref. View  & 86.0 & 66.7  & 68.2 & \textbf{84.3} & 67.7 & 75.1 & \textbf{100.0} & 65.4 & 79.1 \\
Base VLM        & 86.0 & 55.7 & 55.1 & 50.3 & 32.3 & 39.3 & 81.4 & 16.4 & 27.2 \\
\bottomrule
\end{tabular*}

\label{tab:basevlm_full_ref_vqa_cap}
\end{table*}

In this section, we compare \ours{}, which selects a compact subset of highly attended visual tokens from a single reference view per concept, against a baseline that provides the entire reference image as in-context input. We also include a \textbf{Base VLM} configuration, where the model is queried without reference images or concept memories, in order to quantify the benefit of personalization over blind predictions by a non-personalized LVLM.

As shown in Table~\ref{tab:basevlm_full_ref_vqa_cap}, \ours consistently surpasses the Full Reference View baseline while using less than one-fifth of the context budget and eliminating the need to repeatedly reprocess the reference view through the LVLM’s vision encoder. \ours only processes a reference view once when the concept is introduced.

On the \textbf{Recognition} task, the Full Reference View achieves perfect precision (\textbf{100\%}) in the multi-concept setting, but suffers from substantially lower recall compared to \ours{} (\textbf{65.4\%} vs. \textbf{78.2\%}). This indicates that \ours{} effectively removes background content that biases the model toward rejecting images where the concept appears in novel environments, resulting in higher recall and F1-score (\textbf{79.1\%} vs. \textbf{88.6\%}). The benefits of reducing noise in concept memory and extracting representative visual tokens become even more evident in more complex tasks such as VQA, where \ours{} consistently surpasses the baseline by a significant margin.

Overall, these results demonstrate that providing full reference views as in-context input not only incurs higher computational overhead but also injects harmful background noise. In contrast, the attention-guided selection used by \ours{} yields a compact yet highly effective concept memory, eliminating irrelevant background information and removing the need to re-encode reference images at inference time.

Finally, the Base VLM results highlight the necessity of personalization. Without access to concept-specific information, the model's performance drops across all tasks. Although the Base VLM retains reasonable accuracy on the Single VQA task—likely by leveraging general visual cues to make an informed guess between the two available options—it performs poorly elsewhere. In particular, on the Multi-concept recognition task, the F1-score drops drastically from \textbf{88.6\%} to \textbf{27.2\%}. These findings confirm that the strong performance of \ours{} does not stem from dataset biases or generic LVLM capabilities, but instead arises directly from the effectiveness of our attention-guided personalization strategy.





\subsection{Performance on smaller models}
Kang et al.~\cite{kang2025online} present a systematic evaluation of in‑context learning across VLM families and scales, demonstrating that while recent high‑capacity VLMs exhibit strong in‑context abilities, tiny models (\textless4B parameters) still lack visual in‑context learning. In contrast, models in the small–medium range (4B–40B) consistently retain this capability. Our findings in Table~\ref{tab:intern_sizes} reflect this trend: for Yo’LLaVA dataset, InternVL3‑14B achieves the strongest performance, the 8B variant remains competitive, and the 2B model shows reduced precision due to increased false positives. Moreover, as discussed in the main paper, \ours assumes a VLM with sufficient in‑context reasoning ability; we argue that, given the availability of capable open‑source models in this size range, older models naturally become less relevant. \ours thus enables training‑free personalization for modern, capable small‑ and medium‑scale VLMs.

\begin{table}
    \centering
    \caption{\textit{Ego}'s performance with various model sizes and reference objects segmented with G-SAM on Yo'LLaVA dataset.}
    \begin{tabular}{lccc}
\toprule
\textbf{InternVL Size} & \textbf{Prec.} $\uparrow$ & \textbf{Rec.} $\uparrow$ & \textbf{F1} $\uparrow$ \\
\midrule
2B (tiny) & 48.2 & \textbf{91.5} & 63.1 \\
8B (small)  & 88.4 & 81.1 & 84.6 \\
14B (medium) & \textbf{85.0} & 86.4 & \textbf{85.7} \\
\midrule
\makecell{14B \& \textbf{G-SAM}}      & 83.2 & 94.3 & 88.4 \\
\bottomrule
\end{tabular}
    \label{tab:intern_sizes}
\end{table}

\subsection{\ours with segmentation masks}
\ours's attention-guided token selection is inherently a soft region localization method. Therefore, in this section we compare it to the scenario where we incorporate Grounded-SAM (G-SAM)~\cite{ren2024grounded} to obtain segmentation masks of reference objects. Tab.~\ref{tab:intern_sizes} (last row) shows that full segmentation masks improve recall and increase F1 on Yo’LLaVA dataset but at the cost of lower precision, reliance on an external model, added compute/memory, and semantic category supervision. As it encodes full‑mask embeddings, it includes many redundant (possibly un‑informative) patches that can exceed in‑context limits when many concepts appear. In contrast, \ours achieves competitive performance without external modules or semantic labels via attention‑guided, dynamic patch selection, keeping concept memory compact and informative.

\section{Prompt Templates}In this section, we detail the prompts used in different parts of our pipeline for personalization and its evaluation. For simplicity, we provide prompts concerning a single personalized object here. However, multi-concept scenarios could be solved by incrementally appending similar prompts for each personalized subject to the query.

\subsection{\ours's Prompts}
As discussed in Section 3.2 of the main paper, for each concept $c$, we identify a minimal subset of visual tokens  $\mathbf{X}_R^{c}$ from the reference views $R_c$ that best represent the concept. The selection process is guided by cross-modal attention between the visual tokens and the model-generated descriptive keywords. The number of selected tokens is defined based on an estimated concept size within each reference view. To obtain the size estimates and keywords, we use the prompt format shown below when querying the LVLM.

\myparagraph{Concept-Size Estimation\label{sec:concept_size}}{}
\begin{minted}[breaklines, fontsize=\small]{text}
Please analyze the image and estimate the percentage of the total image area that the main subject occupies. If you can not answer, say 0%. Answer **only** the percentage. Answer example: 50%.
\end{minted}
\myparagraph{Keywords Generation}{}
\begin{minted}[breaklines, fontsize=\small]{text}
Give me a list of important words to describe the **main** subject of the image (e.g. blue wheels, green eyes, zigzag pattern, tinted windows...). Provide the list in this exact format: <characteristic 0>, <characteristic 1>, <characteristic 2>,... . Do not answer anything else than the list of important words. Do not mention anything about the background or other objects in the image.
\end{minted}
\myparagraph{In-context Prompting}
As mentioned in section 3.5 of the main paper, at inference, we retrieve the memories $\mathbf{X}_R^{c}$ of the personalized concepts and inject them into the context of the LLM as soft-prompts.
\begin{minted}[breaklines, fontsize=\small, escapeinside=||]{text}
Image <i> shows the entity <c>. Image <i>: |$\mathbf{X}_R^{c}$|.
\end{minted}

Depending on the task the in-context prompt might include one or more concepts from the concept set $C$. $<i>$ denotes each concept's corresponding index in the personalized object list containing $I$ concepts in total. 

\myparagraph{Recognition
}
Given that LVLMs exhibit varying sensitivities to prompt formulations, usually instruction prompts are specific for each model and task. Since our method is training-free, a specific prompt is crafted based on the base model's behavior. We noticed that for more capable models, as in the case of InternVl3, a simple instruction prompt is sufficient. However, we observed that QWEN2.5-VL is trained to abstain from saying yes on people, for this, we adapted the prompt to allow for first probability estimation and then instruct the model to provide a final answer based on the estimated probability of object presence.

\textbf{InternVL3}
\begin{minted}[breaklines, fontsize=\small]{text}
Focusing on each subject's distinctive features, check the presence of the subjects **one by one** in the **new** {media}. Answer with the following template: subject_name: yes/no.
\end{minted}
Where the \{media\} placeholder could be any of `Video' or `Image'.

\textbf{Qwen2.5-VL}
\begin{minted}[breaklines, fontsize=\small]{text}
Do you see any entity in IMAGE <I+1> that resemble <c>? It can appear in a different context, pose or size in IMAGE <N+1>. Answer with a similarity score between 0 - 100. If the similarity score is lower than 50, give the Final Answer as 'No', otherwise as 'Yes'. Your answer should follow this format: Similarity Score: [0-100] Final Answer: [Yes/No].
\end{minted}
\myparagraph{VQA}
For the VQA task we simply append the question to the prompt:
\begin{minted}[breaklines, fontsize=\small]{text}
Answer the following question about Image <I+1>: {question}
\end{minted}
\myparagraph{Captioning}
For the captioning task we use the following prompt:
\begin{minted}[breaklines, fontsize=\small]{text}
Generate a detailed caption describing what you see in Image <I+1>. If an entity was detected, include its given name in the caption.
\end{minted}
\subsection{Autograding with GPT}
As explained in Section 4.1 of the main paper, we adapt an autograding mechanism via GPT 3.5, introduced by \cite{maaz2024video} and used by \cite{seifi2025personalization}, to measure the open-ended VQA performance of different methods. Specifically, we use the following prompt template:
\begin{minted}[breaklines, fontsize=\small]{text}
You are an intelligent chatbot designed for evaluating the correctness of generative outputs for questioqualitatinqualitati-answer pairs. Your task is to compare the predicted answer with the correct answer and determine if they match meaningfully. Here's how you can accomplish the task:
INSTRUCTIONS: 
- Focus on the meaningful match between the predicted answer and the correct answer.
- Consider synonyms or paraphrases as valid matches.
- Evaluate the correctness of the prediction compared to the answer.
Please evaluate the following question-answer pair:
Question: {question}
Correct Answer: {answer}
Predicted Answer: {pred}
Provide your evaluation only as a Yes/No.
DO NOT PROVIDE ANY OTHER OUTPUT TEXT OR EXPLANATION.
\end{minted}
\section{Qualitative Results}

\begin{figure*}[htbp]
      \centering
      \includegraphics[width=\textwidth]{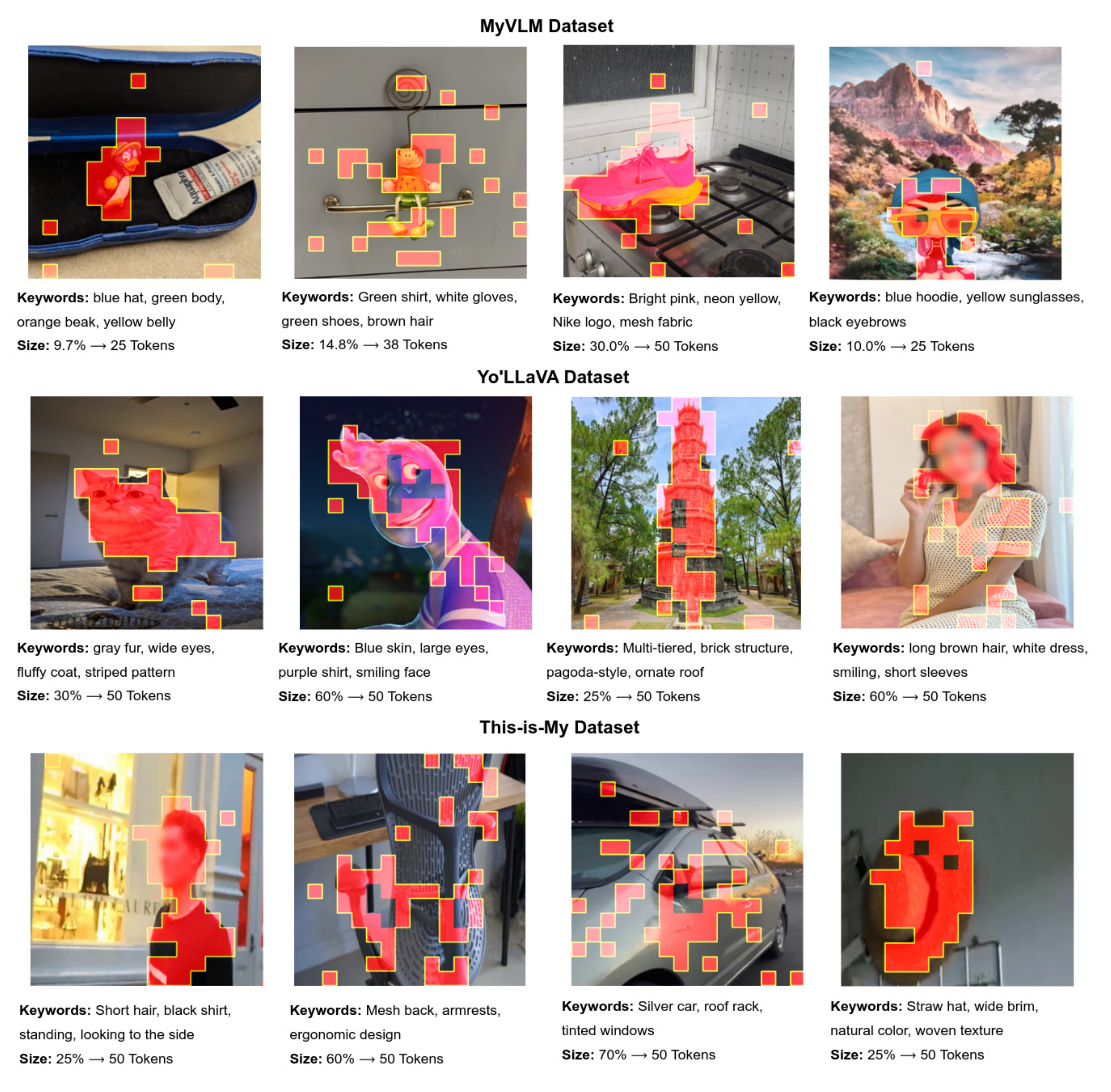}
      \caption{
\ours's generated keywords, estimated concept sizes, and selected patches are shown using examples from various datasets. \ours demonstrates the ability to accurately estimate concept sizes and extract informative patches for each object while minimizing background interference.}
      \label{fig:supp_qualitative}

\end{figure*}

In this section, we provide additional qualitative ablations of our method. Figure~\ref{fig:supp_qualitative} visualizes the visual tokens extracted by \ours across multiple personalization datasets, overlaid on the corresponding reference images. As shown, \ours reliably localizes the target concepts and provides accurate estimates of their spatial extent. However, as noted in Section~\ref{supp:size_selection} of the appendix, objects in the reference views often occupy a large portion of the image. To maintain efficiency and avoid excessive redundancy, we therefore cap the number of extracted visual tokens per reference view to 50.

\begin{figure*}[htbp]
      \centering
      \includegraphics[width=\textwidth]{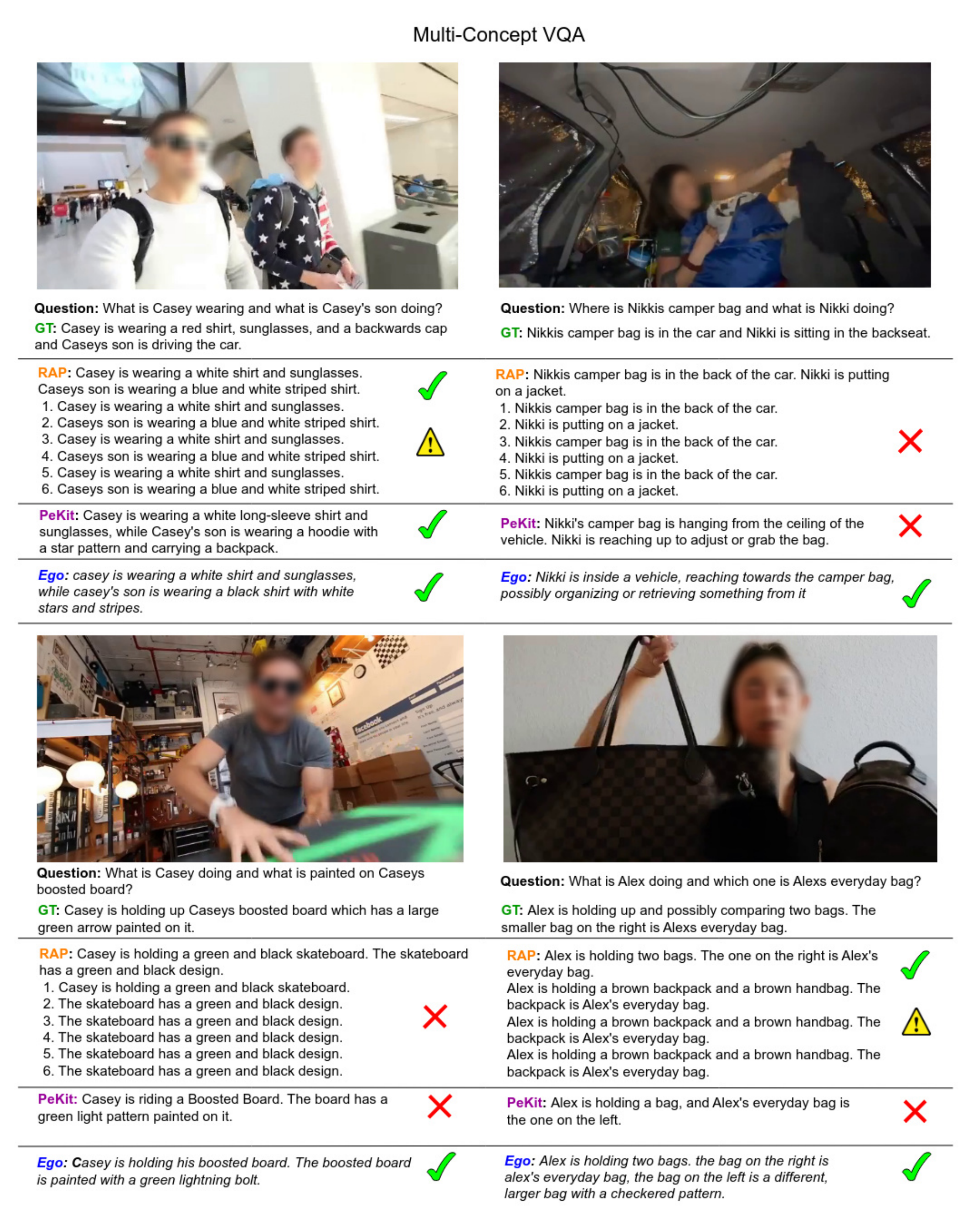}
      \caption{Qualitative results on multi-concept VQA in the This-is-My dataset. \ours maintains accurate responses, while RAP~\cite{hao2025rap} modifies the model’s behavior and PeKit~\cite{seifi2025personalization} fails on fine-grained distinctions.}
      \vspace{-0.1cm}
\label{fig:supp_qualitative_vqa}
\end{figure*}

Figure~\ref{fig:supp_qualitative_vqa} also compares \ours to existing personalization approaches on the multi-concept split from the This-is-My dataset. RAP’s training-based personalization strategy~\cite{hao2025rap} can degrade the base VLM’s generation behavior, occasionally producing incorrect or redundant text. PeKit~\cite{seifi2025personalization} relies on a fixed similarity threshold for instance identification, which is sub-optimal for distinguishing intra-category instances (e.g., Alex’s bag), and its bounding-box overlays may introduce visual biases that lead to hallucinated answers. In contrast, \ours consistently produces correct responses, even in cases where prior methods fail.


\end{appendices}


\end{document}